\documentclass[10pt,journal,compsoc]{IEEEtran}
\ifCLASSOPTIONcompsoc
  \usepackage[nocompress]{cite}
\else
  \usepackage{cite}
\fi
\ifCLASSINFOpdf
  \usepackage[pdftex]{graphicx}
  \graphicspath{{Fig/}}
\else
\fi
\usepackage[cmex10]{amsmath}
\usepackage{bm}
\DeclareMathOperator*{\argmin}{arg\,min}
\usepackage{algorithm}
\usepackage{algorithmic}

\usepackage{booktabs}
\usepackage{multirow}
\usepackage{color}

\ifCLASSOPTIONcompsoc
 \usepackage[caption=false,font=footnotesize,labelfont=sf,textfont=sf]{subfig}
\else
 \usepackage[caption=false,font=footnotesize]{subfig}
\fi
\usepackage{amssymb,changes}
\usepackage{soul}
\begin{document}
%
\title{Cascaded Regression using Landmark Displacement for 3D Face Reconstruction}
%
%
%
%

\author{~Fanzi~Wu,
        Songnan~Li,
        Tianhao~Zhao,
        and~King~Ngi~Ngan,
        Lv~Sheng
}

%
%

\markboth{submitted to pattern recognition letters}%
{Shell \MakeLowercase{\textit{et al.}}: Bare Demo of IEEEtran.cls for Computer Society Journals}
%



\IEEEtitleabstractindextext{%
\begin{abstract}
This paper proposes a novel model fitting algorithm for 3D facial expression reconstruction from a single image. Face expression reconstruction from a single image is a challenging task in computer vision. Most state-of-the-art methods fit the input image to a 3D Morphable Model (3DMM). These methods need to solve a stochastic problem and cannot deal with expression and pose variations. To solve this problem, we adopt a 3D face expression model and use a combined feature which is robust to scale, rotation and different lighting conditions. The proposed method applies a cascaded regression framework to estimate parameters for the 3DMM. 2D landmarks are detected and used to initialize the 3D shape and mapping matrices. In each iteration, residues between the current 3DMM parameters and the ground truth are estimated and then used to update the 3D shapes. The mapping matrices are also calculated based on the updated shapes and 2D landmarks. HOG features of the local patches and displacements between 3D landmark projections and 2D landmarks are exploited. Compared with existing methods, the proposed method is robust to expression and pose changes and can reconstruct higher fidelity 3D face shape. 
\end{abstract}

}

\maketitle

\IEEEdisplaynontitleabstractindextext

%
\IEEEpeerreviewmaketitle

\IEEEraisesectionheading{\section{Introduction}\label{sec:introduction}}

%
%
%
%
\IEEEPARstart{T}{he} facial expression is important in the face analysis since it conveys our emotions, thoughts and intentions. Human perception is sensitive to expression changes, even subtle ones. Therefore, 3D face expression reconstruction from images is widely investigated as an important computer vision task. Face geometry reconstruction from a single RGB image is an ill-posed problem due to self-occlusion, various lighting conditions, facial textures and so on, thus particularly challenging. To cope with these difficulties, various priors for 3D face reconstruction have been proposed, and the most widely used one is the 3D morphable model (3DMM)\cite{blanz1999morphable}. The 3DMM converts the target from a 3D object to a limited number of 3DMM parameters, which reduces the dimension of the output significantly. Analysis by synthesis methods fit the 3DMM to the input image by optimizing a tailored objective function\cite{blanz2003face}\cite{romdhani2003efficient}\cite{romdhani2005estimating}. However, these methods need to carefully design the regularization terms and iteratively solve a nonlinear optimization problem which is usually time-consuming. Recently, learning-based methods are proposed to directly learn a regressor from training data to estimate model parameters\cite{cao2015real}\cite{zhu2015discriminative}. 3D face datasets\cite{yin2008high}\cite{phillips2005overview} usually do not provide corresponding model parameters, therefore, these methods need to prepare model parameters for training. There is no standard algorithm for data preparation which makes the learning methods incomparable. 


To establish fair comparison, we elaborated a 3D face dataset with 3DMM parameters and 3D meshes. The 3DMM parameters are used for training and the 3D meshes has sample correspondence with 3MM, which could be used as ground truth for evaluation. Based on such dataset, we proposed a cascaded regression method which used a high level feature to represent facial attributes. In the training stage, landmarks fitting is used to initialize the 3DMM parameters and the mapping matrices before the regression. A joint feature which has patch information and semantic meaning is used. To eliminate the limitation of training data, mapping matrices are calculated instead of regressing. The proposed method can reconstruct the face geometry with high accuracy and robustness to various expressions and poses.  

 The main contribution of this paper is three-fold:
 \begin{itemize}
 \item we propose a 3D face reconstruction framework which is robust to expression and pose changes;
 \item We propose a high level feature - Landmark Displacement (LD), which has semantic meaning and robustness to lighting variations; 
 \item we provide 3DMM parameters and 3D meshes for a 3D face expression database which now can be used to develop and evaluate learning-based 3D face reconstruction algorithms.
 \end{itemize}
 
This paper is organized as follows: In Section 2, we briefly introduce the 3DMM and related model fitting methods. In Section 3, the proposed method is elaborated in several processing steps including landmark fitting, feature extraction and regression. In Section 4, we explain the experimental setup and data preparation, and compare with state-of-the-art methods. In the last section, we draw the conclusion and discuss the future work.
 
\section{Related Work}
 \subsection{3D Morphable Model} 
 3DMM is a statistical model derived from registered 3D face scans\cite{blanz1999morphable}. The most widely used 3DMM is the Basel Face Model (BFM)\cite{paysan20093d} proposed by Paysan \MakeLowercase{\textit{et al.}}. A registered 3D face scan has two major components: geometry $S = [X_1,Y_1,Z_1,...,X_n,Y_n,Z_n]$ and texture $T = [R_1,G_1,B_1,...,R_n,G_n,B_n]$, which represent the 3D location and RGB color, respectively, for each vertex. Principal Component Analysis(PCA) is applied to the matrix consisting of geometry $S$ or texture $T$ of all face scans. The eigenvectors can be used to synthesize different shapes and textures: 
\begin{subequations}
\label{eq2_1}
\begin{align}
\bm{S}=\overline{\bm{S}}+\sum\bm{\alpha}_i\bm{s}_i \\
\bm{T} = \overline{\bm{T}}+\sum\bm{\gamma}_i\bm{t}_i
\end{align}
\end{subequations}
where $\bm{\alpha}$ and $\bm{\gamma}$ are parameters for shape and texture; $\bm{s}_i$ and $\bm{t}_i$ are their corresponding eigenvectors. 3DMM provides a strong regularization and converts the optimization of $3\times n$ (typically greater than $10000$) variables to a much lower dimension (around $100$). However, BFM is incapable of sythesizing facial expressions. Cao \MakeLowercase{\textit{et al.}} \cite{cao2014facewarehouse} used a bilinear face model which can synthesize faces with different identities and expressions using 
\begin{equation}
  \bm{S} = \bm{C}_r \times_2 \bm{\alpha}_{id} \times_3 \bm{\beta}_{exp}
\label{eq2_2}
\end{equation}
where $\bm{C}_r$ is a reduced core tensor provided by the FaceWarehouse database; operation $\times_n$ is the mode-$n$ multiplication; $\bm{\alpha}_{id}$ and $\bm{\beta}_{exp}$ are the identity and expression parameters, the dimensions of which are chosen to be 50 and 25, respectively, in our experiments. The reconstruct shape $S$ contains $x,y,z$ coordinate values for all vertices.
  
  To get the mapping from 3D to image coordinate, either perspective or orthographic projection can be used. For efficient optimization and also fair experimental comparison with other methods, orthographic projection are chosen in this work. 
  For the $i_{th}$ vertex,
  \begin{equation}
   \label{eq2_3}
   \bm{v}_i = \bm{C}_{r,i}\times_2 \bm{\alpha}_{id} \times_3 \bm{\beta}_{exp}
  \end{equation}
  $\bm{C}_{r,i}$ is a portion of the reduced core tensor associated with the $i_{th}$ vertex. The orthographic projection of $\bm{v}_i$ is:
  \begin{equation}
   \label{eq2_4}
   \bm{p} = \bm{A}(\bm{R}\bm{v}_i + \bm{t})
  \end{equation}
  or equivalently 
  \begin{equation}
   \label{eq2_5}
   \begin{bmatrix}
   x\\y
   \end{bmatrix}
     = 
    \begin{bmatrix}
    a&0&0\\
    0&a&0
    \end{bmatrix}
    \begin{bmatrix}
     r_{11}&r_{12}&r_{13}\\
     r_{21}&r_{22}&r_{23}\\
     r_{31}&r_{32}&r_{33}\\
    \end{bmatrix}
    \begin{bmatrix}
     X\\Y\\Z
    \end{bmatrix}
    +
    \begin{bmatrix}
    t_1\\t_2
    \end{bmatrix}
  \end{equation}
  where $\bm{p}$ is the projection vector in the image coordinate; $\bm{A}$ is a scaled orthographic projection matrix; $\bm{R}$ is the $3\times3$ rotation matrix and $\bm{T}$ is the $2\times1$ translation vector. The proposed method estimates two groups of variables: the 3DMM parameters $\bm{g}=\left[\bm{\alpha};\bm{\beta}\right]$ and the mapping matrices $\bm{c} =\lbrace \bm{A},\bm{R},\bm{T}\rbrace$.
  
 \subsection{Model Fitting}
The model fitting process estimates the correspondences between the 2D image and the 3DMM, then minimizes their differences. A popular approach is to find a pixel-by-pixel dense correspondence between the 3DMM and the input image. Various features such as optical flow, SIFT feature, pixel intensity and contours have been investigated\cite{romdhani2005estimating}\cite{blanz2003face}\cite{Fanzi2016sift}. These methods can generate high quality results but are computationally expensive. Romdhani \MakeLowercase{\textit{et al.}} proposed a linear approach to compute an incremental update to the shape and texture parameters given dense measurements of residual errors from optical flow\cite{romdhani2002face}. Besides dense correspondences, Moghaddam \MakeLowercase{\textit{et al.}} used silhouettes computed from a large number of input images which is robust to lighting variations\cite{moghaddam2003model}. Since pixel intensities of a face image can be dramatically influenced by expression, occlusion and complicated illumination, the model fitting result may vary a lot given different imaging conditions. Landmark is another frequently used feature which provides sparse correspondences\cite{aldrian2010linear}. However, in previous work, landmarks are often manually labeled which is inefficient and labor intensive. According to the investigation in \cite{bulat2017far}, state-of-the-art landmark detection algorithms can achieve human-level performance. Thus, the proposed method directly takes the landmark detection results as the input, so as to fully take advantage of the fast progress in this field.
 
Recently machine learning are widely used in computer vision tasks especially for face analysis\cite{cao2014displaced}\cite{Cao2013siggraph}. Different from analysis by synthesis methods which can be easily trapped in a local minimum, learning-based or deterministic methods are more robust and tend to be able to find a global optimum solution, when the training data is sufficient. Zhu \MakeLowercase{\textit{et al.}} proposed to learn a cascaded regression method using HOG features to estimate 3DMM parameter residues\cite{zhu2015discriminative}. This work is one of the pioneers in the learning-based 3D face reconstruction. However, it only worked for neutral frontal faces.  Inspired by \cite{zhu2015discriminative}, we propose a cascaded regression method based on the face expression model and a joint feature which is robust to pose and expression changes. Besides above work, as the development of deep neural network, there are efforts paid on such method. However, as we all known, such deep neural network are data-hungry and relies heavily on the GPU performance. We compared the proposed method with state-of-the-art methods which used deep neural network to regress 3DMM parameters and mapping matrices, and we have comparable results with them while using less training data. 
 
For learning-based approach, training data needs to be prepared, which usually is labor-consuming. To our best knowledge, there is no 3D face database that provides ground truth of 3DMM parameters and 3D meshes. To facilitate algorithm development and performance evaluation, we develop a method to generate 3DMM parameters on the Bosphorus dataset, and release them to the public for research use at \url{http://www.ee.cuhk.edu.hk/~fzwu/projects/BDpar.html}. 
\section{Method}
 \subsection{Overview}
 \begin{figure*}[ht]
 \centering
  \includegraphics[width = 0.9\linewidth]{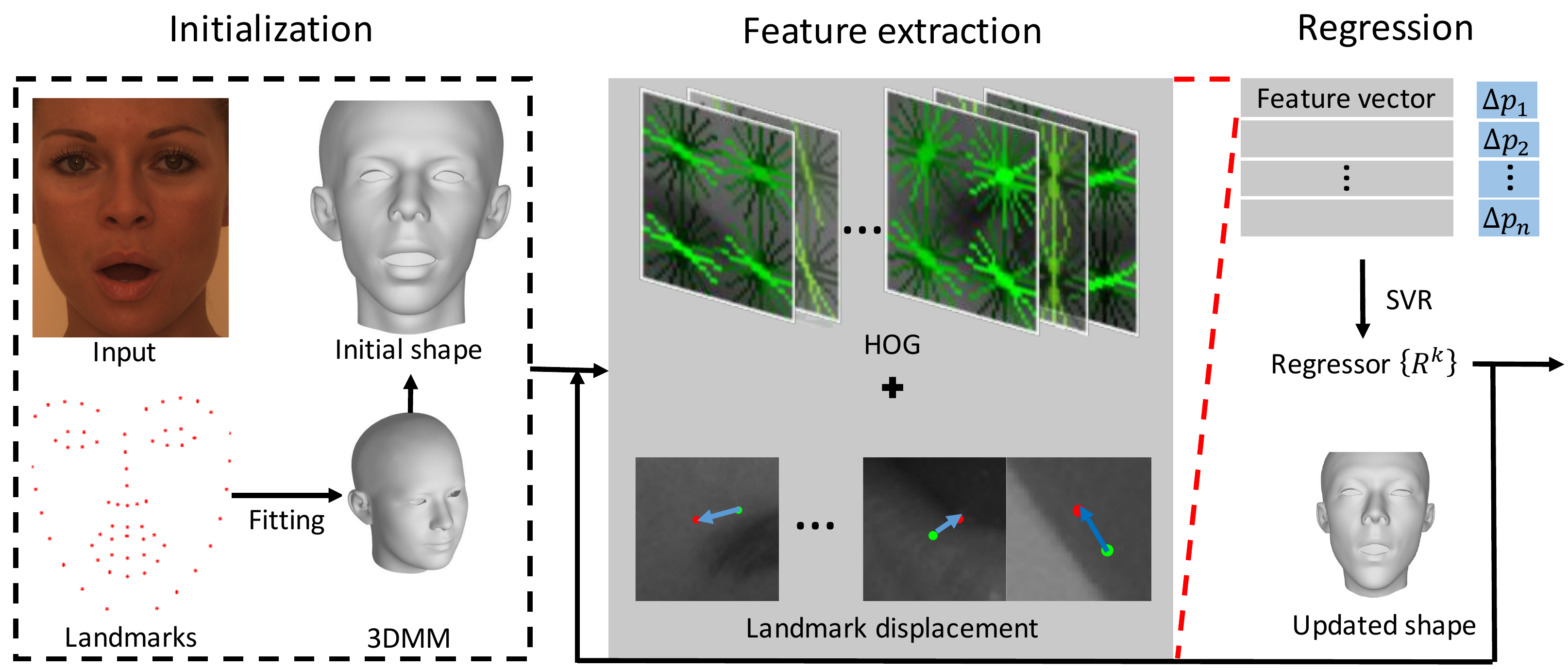}
  \caption{Flowchart of the proposed method. $\triangle p$ is the residue between current parameters and ground truth parameters for each sample. $R^k$ is the regressor learned in the $k^{th}$ iteration.}
  \label{fig1}
 \end{figure*}
 
The flowchart of the proposed method is shown in Fig. \ref{fig1}. For each training sample, model parameters and mapping matrices are initialized using 2D landmarks. A cascade regressor is learned to estimate the residues between initial parameters and the ground truth parameters. A joint feature, which is the combination of HOG and LD is extracted. The regressor is learned using feature vectors and parameter residues $\triangle p$ of all training samples. Then the model parameters for each sample are updated and the mapping matrices are further updated. The next iteration begins with the updated 3D shape and mapping matrices. This updating process iterates until its converge. 
  
 \subsection{Landmarks fitting}
  For each input image $\bm{I}$, 49 inner landmarks and 17 boundary landmarks $\bm{l}$ are detected using Dlib, respectively. Then 2D landmarks are fitted to the 3DMM to estimate model parameters $\bm{\alpha}$, $\bm{\beta}$ and the mapping matrices $c$. Firstly, $c$ is solved by fixing the model parameters to the mean value:
  \begin{equation}
   \label{eq3_7}
   \bm{c} = solver(\bm{l},\bm{C}_{r,i\in L}\times_2 \bm{\mu}_\alpha \times_3 \bm{\mu}_\beta)
  \end{equation}
  where $\bm{l}$ is the 2D landmarks in the image coordinate; $i\in L$ denotes landmark vertices; $\bm{\mu}_\alpha$ and $\bm{\mu}_\beta$ are the mean parameters for identity and expression, respectively; $solver(\cdot)$ is a linear solver introduced in \cite{bruckstein1999optimum}. Then the identity $\bm{\alpha}$ and expression $\bm{\beta}$ are estimated based on $c$ using
   \begin{multline}
    \label{eq3_8}
   \min_{\bm{\alpha},\bm{\beta}}\sum_{i\in L}\Vert l_i - P_i (\bm{\alpha},\bm{\beta},\bm{c})\Vert
   +\lambda_1 (\bm{\alpha} - \bm{\mu}_\alpha)\top \bm{Q}_\alpha(\bm{\alpha} - \bm{\mu}_\alpha)\\
   +\lambda_2  (\bm{\beta} - \bm{\mu}_\beta)\top \bm{Q}_\beta(\bm{\beta} - \bm{\mu}_\beta) 
   \end{multline}
  where $P_i(\cdot)$ calculates the projection of the $i_{th}$ 3D landmark; $\bm{Q}_\alpha$ and $\bm{Q}_\beta$ are diagonal matrices which contain the reciprocal of the variance of each identity and expression parameter, respectively. The regularization term in Eq. (\ref{eq3_8}) penalizes deviation from the mean, with $\lambda_1$ and $\lambda_2$ controlling the regularization intensity. This processing step provides an initialization for the feature extraction and helps the regression process converge faster.
  
 \subsection{Feature extraction}
 Feature extraction is a critical part in learning-based methods. Good features can fully represent the facial information from the input image and is robust to  Different features has been used, such as HOG, SIFT and LBP. However, those features are based on image intensity, gradient and has no semantic meaning. In this paper, we proposed LD feature, which represents the displacement between points. This feature utilize the semantic meaning of facial landmarks and can better represent facial attribute. 
 
 In each regression stage, he HOG features are extracted from $64\times 64$ local patches of the input image $I$ around the 3D landmark projections. The LD feature is defined as the difference between each landmark position $l_i$ and the projection position of its corresponding model vertex $p_i$:
  \begin{equation}
   \label{eq3_9}
   \bm{u} = [\bm{u}_1, \bm{u}_2, ..., \bm{u}_L] \\
  \end{equation}
  where
  \begin{equation}
   \label{eq3_9b}
   \bm{u}_i = \bm{l}_i - P_i(\bm{g},\bm{c})
  \end{equation}
HOG and LD features are concatenated together as:  
\begin{equation}
  \mathcal{F} = 
  \left[
  \begin{matrix}
   \bm{h}\\
   \bm{u}
  \end{matrix}
  \right]=
  \left[
  \begin{matrix}
   \bm{h}_1& \bm{h}_2& ...&\bm{h}_L\\
   \bm{u}_1& \bm{u}_2& ...&\bm{u}_L 
  \end{matrix}
  \right]
 \end{equation}
 where $\bm{h}_L$ is the hog feature vector for each landmark.
The LD feature uses the strong correspondences between 3D and 2D landmarks and is robust to lighting variations. State-of-the-art landmark detection algorithms already can achieve human-level performance especially for frontal face images\cite{bulat2017far}. Therefore in this work, we simply assume the detected landmarks are reliable (which is true for most images in our experiments) and can be exploited for 3D face reconstruction. On the other hand, the HOG feature uses pixel colors of local image patches, which contain rich information about the facial albedo, geometry, and lighting condition. The joint feature of HOG and LD fully takes advantages of the input data, and robust to lighting, scale and rotation variations.
  
 \subsection{Regression}
 Currently there are only a few 3D databases that provide large pose face scans which makes the training for large pose cases difficult. To alleviate this problem, we estimate the model parameters using regression, but calculate the mapping matrix using a linear  solver as given in Eq. (\ref{eq3_7}). We combine Eqs. (\ref{eq2_3})(\ref{eq2_4}) into a function $P(\bm{g},\bm{c})$ to calculate the orthographic projection of the 3D shape $\bm{S}$. Xiong \MakeLowercase{\textit{et al.}} proposed the Supervised Descent Method (SDM) \cite{Xiong:2013wv}, in which the regressor estimates the incremental value $\triangle \bm{g}$ instead of parameters $\bm{g}$, as given below
  \begin{equation}
   \label{eq3_5}
   \min_{\triangle \bm{g}}\Vert \bm{g}^\star - (\bm{g} + \triangle \bm{g})\Vert
  \end{equation}
  where $\bm{g}^\star$ denotes the ground truth and $\triangle \bm{g}$ is the incremental value
  \begin{equation}
   \label{eq3_6}
   \triangle \bm{g} = \bm{\mathcal{R}}(\mathcal{F}(\bm{I},P_{i\in L}(\bm{g},\bm{c})))
  \end{equation}
  In Eq. (\ref{eq3_6}), $\mathcal{F}$ extracts features at color patches of the input image $I$ which center around the projections of 3D landmarks $P_{i\in L}(\bm{g},\bm{c})$, and $\bm{\mathcal{R}}$ is the regressor. 
 In each regression stage, both model parameters $g$ and the mapping matrices $c$ are updated.
In the $k_{th}$ stage, $\bm{g}^k$ is updated using the learned regressor $\bm{\mathcal{R}}^k$.
  \begin{equation}
   \label{eq3_10}
   \bm{\mathcal{R}}^k = \argmin_{\bm{\mathcal{R}}^k} \sum^N_{j=1}\Vert (\bm{g}^\star_j - \bm{g}^{k-1}_j)-\bm{\mathcal{R}}^k(\mathcal{F}(\bm{I}_j,P_{i\in L}(\bm{g}^{k-1}_j,\bm{c}_j))) \Vert
  \end{equation}
 Algorithm \ref{alg1} shows the overall training process.

\begin{algorithm}[ht]
   \caption{Learning Combined Cascaded Regressors}
  \begin{algorithmic}[1]
   \label{alg1}
   \REQUIRE Training data $\lbrace (\bm{I}_j,\bm{g}^\star_j)|j=1,2,...,N\rbrace$, landmarks $\bm{l}_j$
   \ENSURE Cascaded Regressor $\lbrace \bm{R}^k\rbrace^K_{k=1}$
   \STATE Initialize $\bm{g}^0_j$, $\bm{c}^0_j$ by Eq. (\ref{eq3_8}) for the $j_{th}$ sample;
   \FOR {$k = 1 \to K$}
    \STATE Extract features $\mathcal{F}_j$ using $\bm{g}^{k-1}_j$, $\bm{c}^{k-1}_j$, and $\bm{l}_j$;
    \STATE Estimate $\bm{R}^k$ via Eq. (\ref{eq3_10});
    \STATE Estimate $\triangle \bm{g}^k_j$ via Eq. (\ref{eq3_6}) and update current parameter by $\bm{g}^k_j = \bm{g}^{k-1}_j + \triangle \bm{g}^k_j$ for the $j_{th}$ sample;
    \STATE Update the mapping matrix $\bm{c}_j$ for the $j_{th}$ sample;
   \ENDFOR
  \end{algorithmic}
  \end{algorithm}

\section{Experiments}
 \subsection{Data Preparation}
 In our experiments, Bosphorus Database was used as the training and testing data\cite{savran2008bosphorus}\cite{SAVRAN2012774}. The Database contains 105 subjects and 4666 facial samples in total. Each subject has six emotional expressions and several head poses over $0-90$ degree. Each sample associates with a 2D image $\bm{I}$, a 3D point cloud $\bm{D}$, and manually-labeled landmarks both in 2D and 3D. Samples which failed in the landmark detection were dropped, since the proposed method relies on the detected landmarks. Usually these failure samples are occluded or have yaw rotation with 90 degree, which are less than $10\%$ of the dataset. We conducted five cross-validation experiments, and in each experiment 100 subjects were used for training and 5 unknown subjects were used for testing. 

\begin{figure}[ht]
\subfloat[]
{\includegraphics[width=0.3\textwidth]{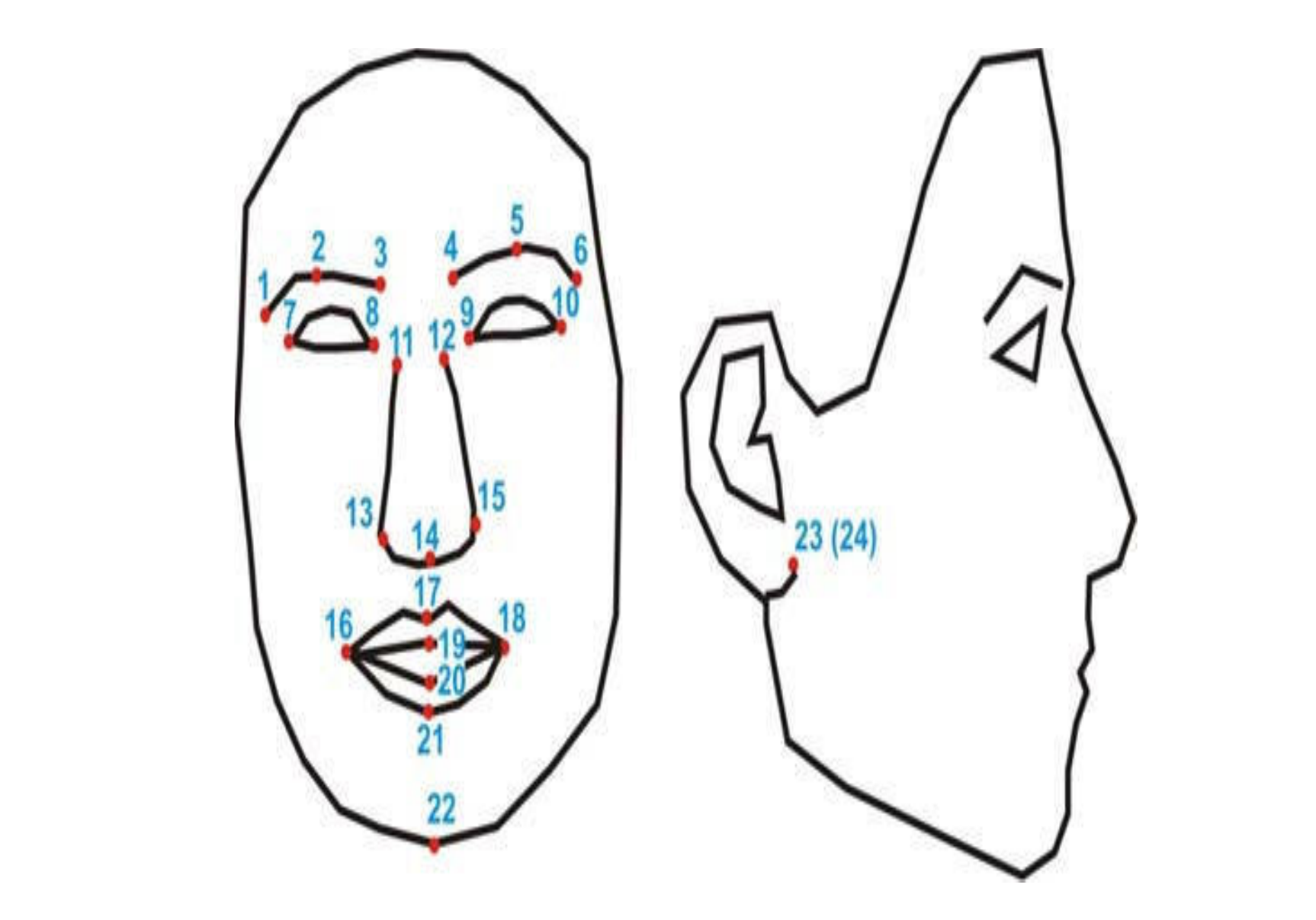}%
\label{fig2_1}}
\subfloat[]
{\includegraphics[width=0.2\textwidth]{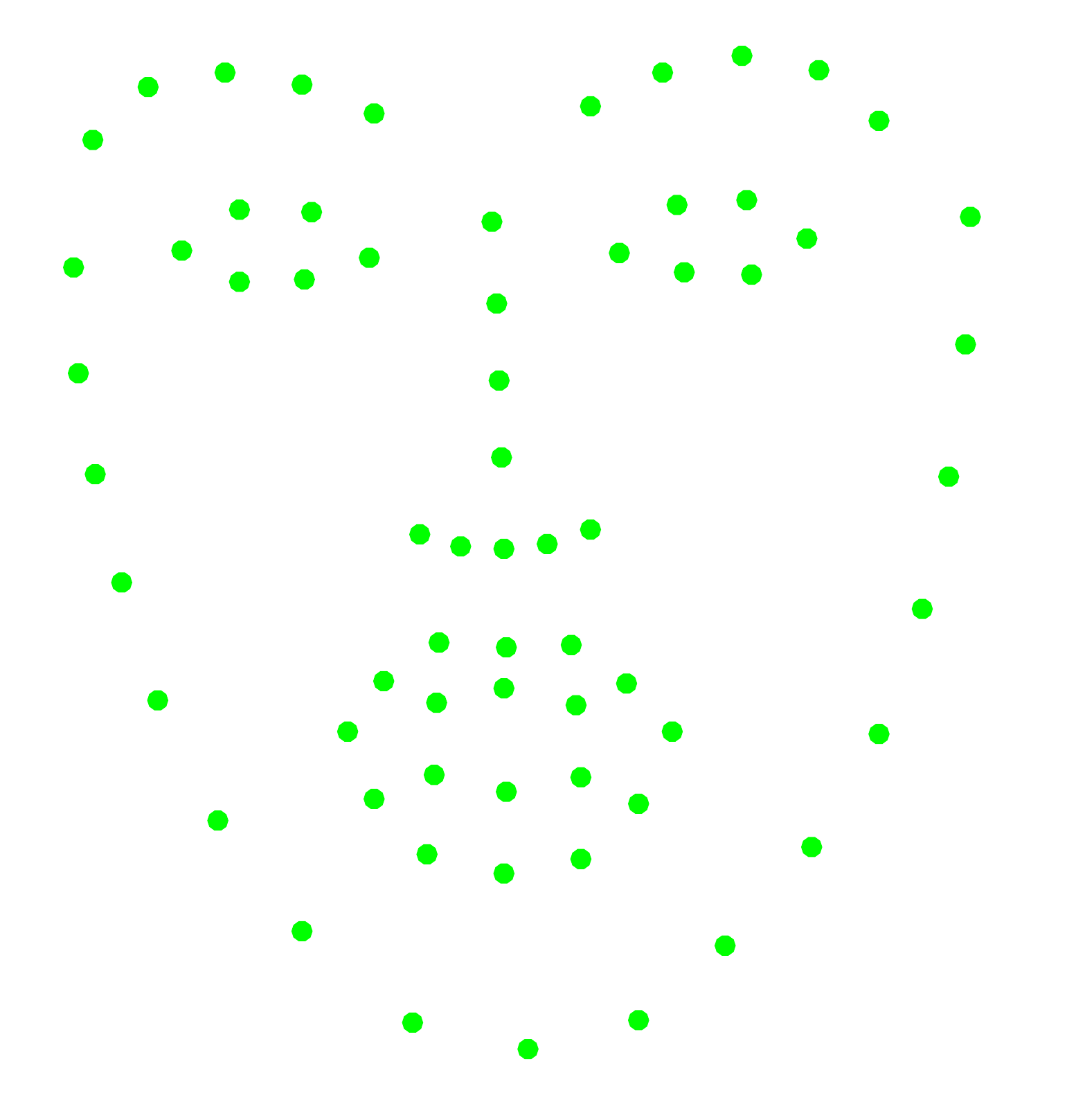}%
\label{fig2_2}}
\caption{Comparison between BD landmark labels and detected landmarks.(a) 24 landmarks of BD (b) 66 landmarks used in the proposed method}
\label{fig2}
\end{figure}

The raw data from Bosphorus Database cannot be used as training data and evaluation ground truth directly. Thus we elaborated Bosphorus Database with 3DMM parameters and 3D meshes registered with 3DMM. Each training pair consists of a RGB image $\bm{I}$ and its associated 3DMM parameters $\bm{g}^\star$. $\bm{g}^\star$ were derived by fitting the 3DMM to the 3D point cloud $\bm{D}$. Firstly, the 3DMM were fitted to the 3D landmarks to initialize the rigid transformation and model parameters. Then, a dense fitting was performed which fitted the 3DMM to both the landmarks and the 3D point cloud. The correspondences and the rigid transformation were updated iteratively using the ICP algorithm. As showed in Fig. \ref{fig2_1}, the manually labeled landmarks provided by the database only include the eye corners, therefore, it cannot deal with cases of closed eyes. To solve this problem,we preprocessed the image using the landmark fitting algorithm introduced in Section 3.2.

Besides, a 3D mesh $M$ that has the same vertex number and semantic meaning as the 3DMM was constructed to evaluate the reconstruction accuracy. It is obtained by performing non-parametric fitting to the 3D point cloud $D$ using the Laplacian deformation method in \cite{sorkine2004laplacian}. 

 \subsection{Evaluation}
 The 3D reconstruction quality can be measured subjectively or objectively. The subjective measurements represent the human perception of the reconstruction quality, while the objective measurements make the comparison between different methods convenient and reproducible. In this paper, two objective quality metrics are used: RMSE and MAE. RMSE evaluates the errors over $z$ coordinate while MAE evaluates the errors over $x$,$y$,$z$ coordinates.
\begin{figure*}[ht]
 \centering 
 \includegraphics[width =0.9\textwidth]{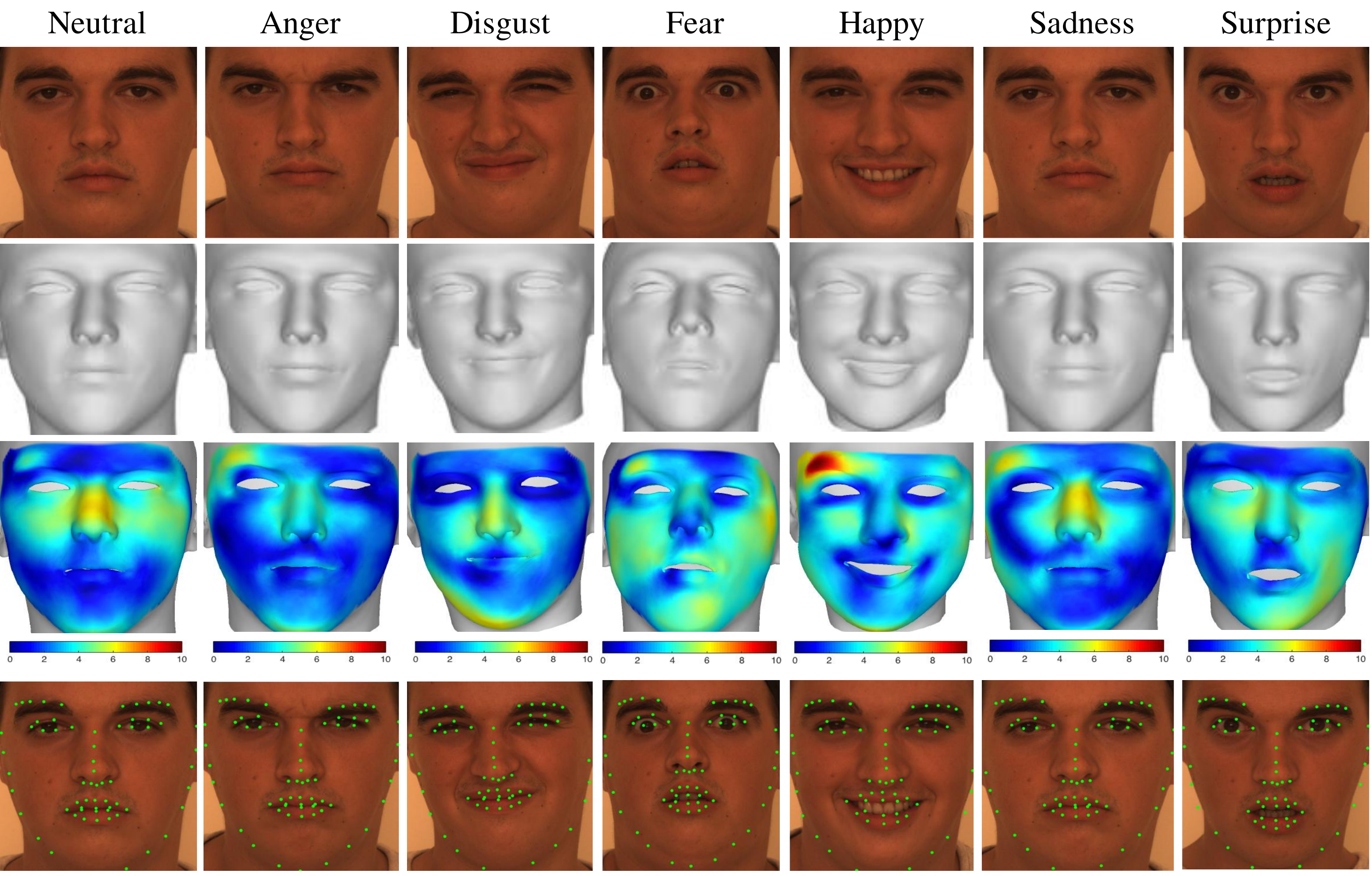}
 \caption{Results of the proposed method over different expressions. Row 1: Input Image. Row 2: Reconstruction results. Row 3: Error map of the MAE between reconstruction results and ground truth meshes. Row 4: Projection of 3D landmarks.}
 \label{fig8}
\end{figure*}
\subsubsection{Root Mean Squared Error(RMSE)} 
The RMSE is defined as
\begin{equation}
RMSE = \sqrt{\frac{\sum_{i=1}^{n}{({z^*_i-z_i})^2}}{n}}
\label{eq4_1}
\end{equation}
where $n$ is the number of vertices, $z^*_i$ and $z_i$ are the $z$ coordinate values for the $i^{th}$ vertex of the ground truth and the reconstructed shape, respectively. 
\subsubsection{Mean Average Error (MAE)} 
MAE is defined as the mean average euclidean distance between the ground truth and the reconstructed shape.
\begin{equation}
MAE = \frac{\sum{||{\bm{v}^*_i-\bm{v}_i}||_2}}{n}
\label{eq4_2}
\end{equation}
where $\bm{v}^*_i$ and $\bm{v}_i$ are the 3D positions for the $i^{th}$ vertex of the ground truth and the reconstructed shape, respectively.


\begin{figure}[ht]
 \centering
 \includegraphics[width =\linewidth]{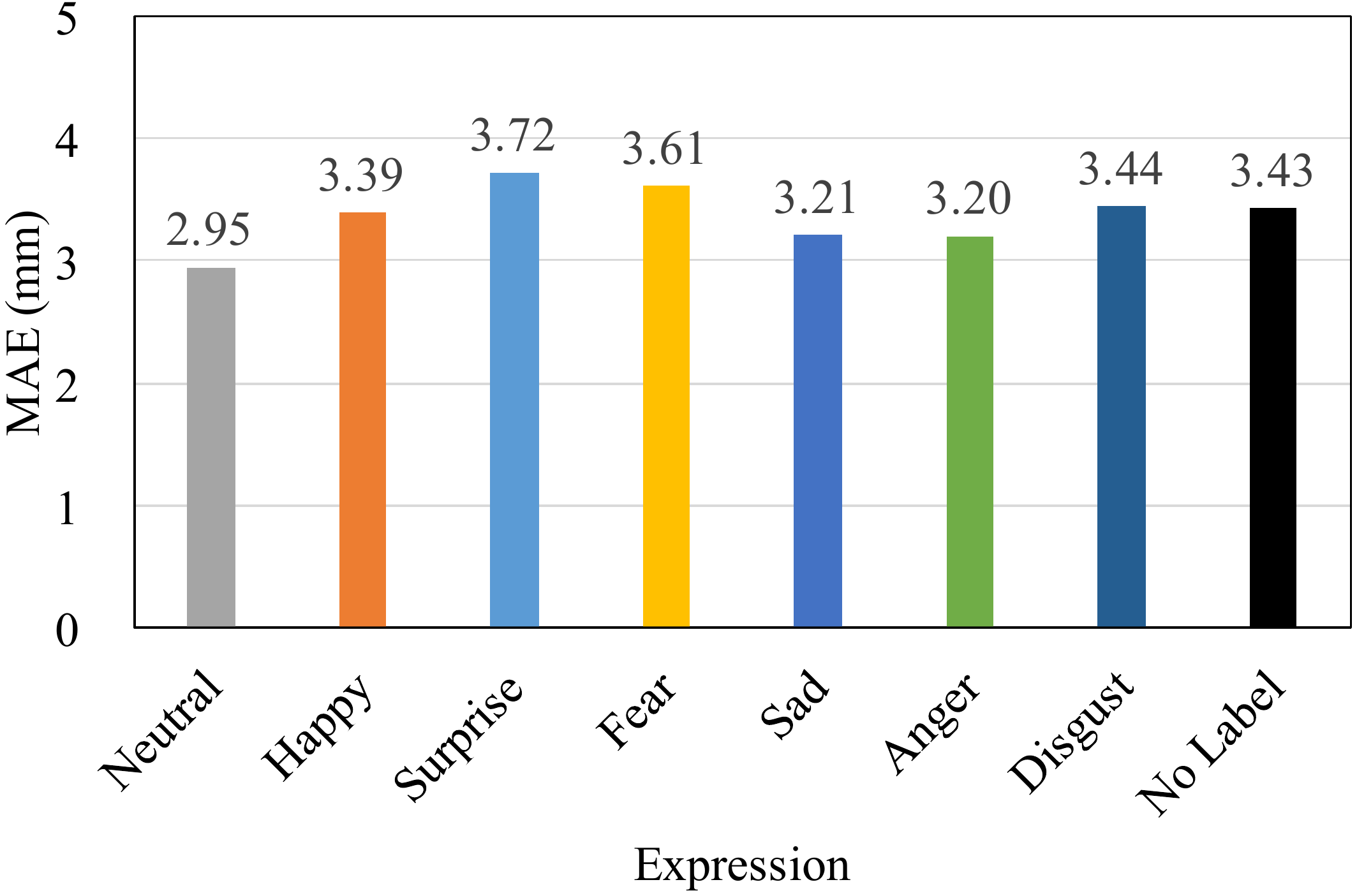}
 \caption{MAE of different expressions. 'No label' has facial motions without emotion meaning.}
 \label{fig7}
\end{figure}

\subsection{Results} 

\subsubsection{Reconstruction accuracy over different expressions} Fig. \ref{fig8} shows the reconstruction results and the error maps over seven expression classes. The color of most area on each error map is blue (less than 4mm) and those error maps have little variation between different expressions. The proposed method reconstructs different expressions accurately and keeps the identity consist. The projection of 3D landmarks also shows that both 3DMM parameters and mapping matrices are accurate. Fig. \ref{fig7} presents the MAE compared with the ground truth mesh $\bm{M}$ across different expressions. The proposed method shows consistent accuracy over different expressions. The maximum difference between tested classes (i.e. Surprise and Neutral) is less than $3\%$.

\subsubsection{Resonctruction accuracy over different poses} 
\begin{figure}[ht]
 \centering
 \includegraphics[width =0.95\linewidth]{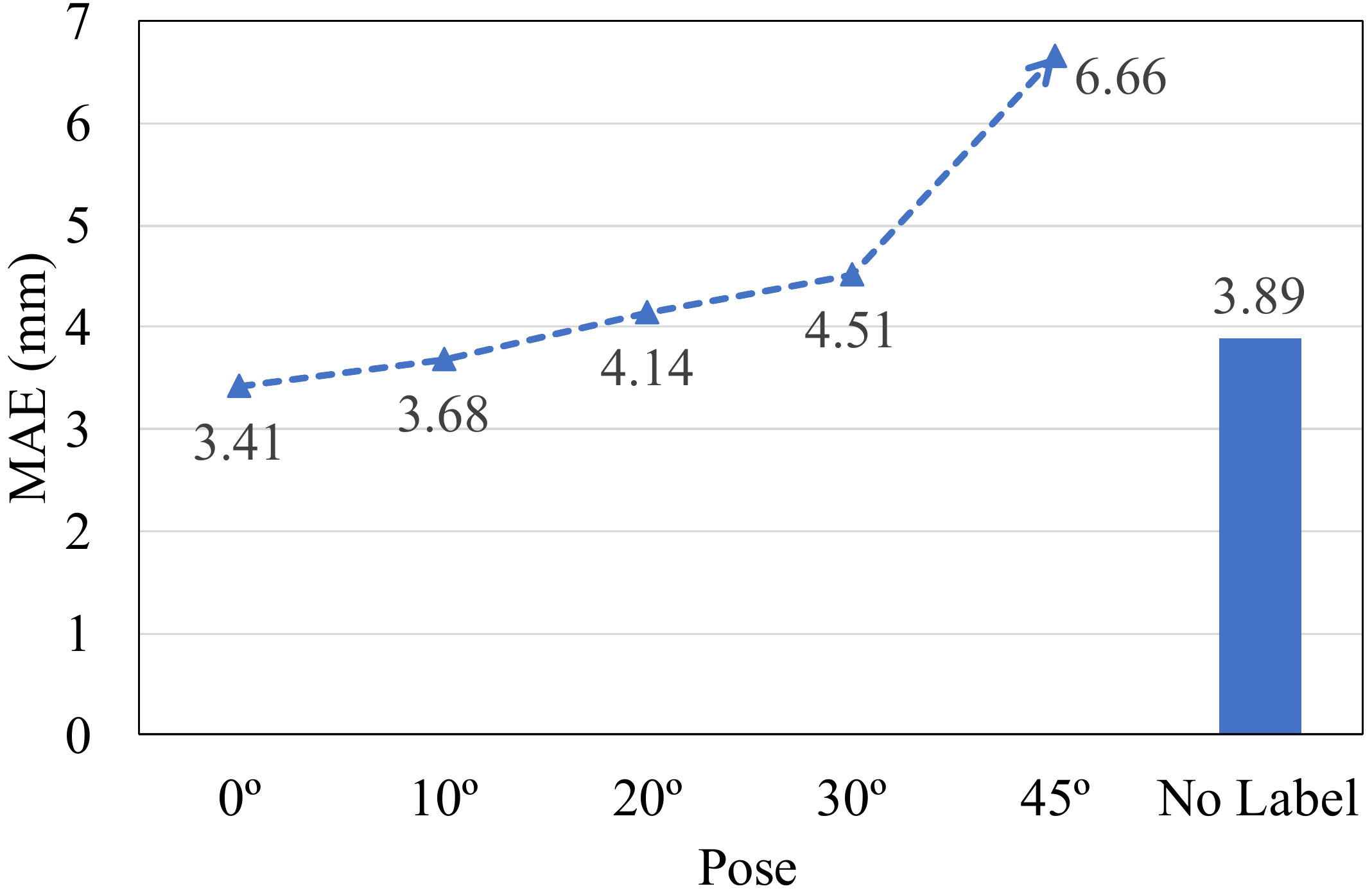}
 \caption{MAE over different yaw poses. 'No label' has both yaw and pitch rotations.}
 \label{fig5}
\end{figure}
\begin{figure}[ht]
 \centering
 \includegraphics[width = \linewidth]{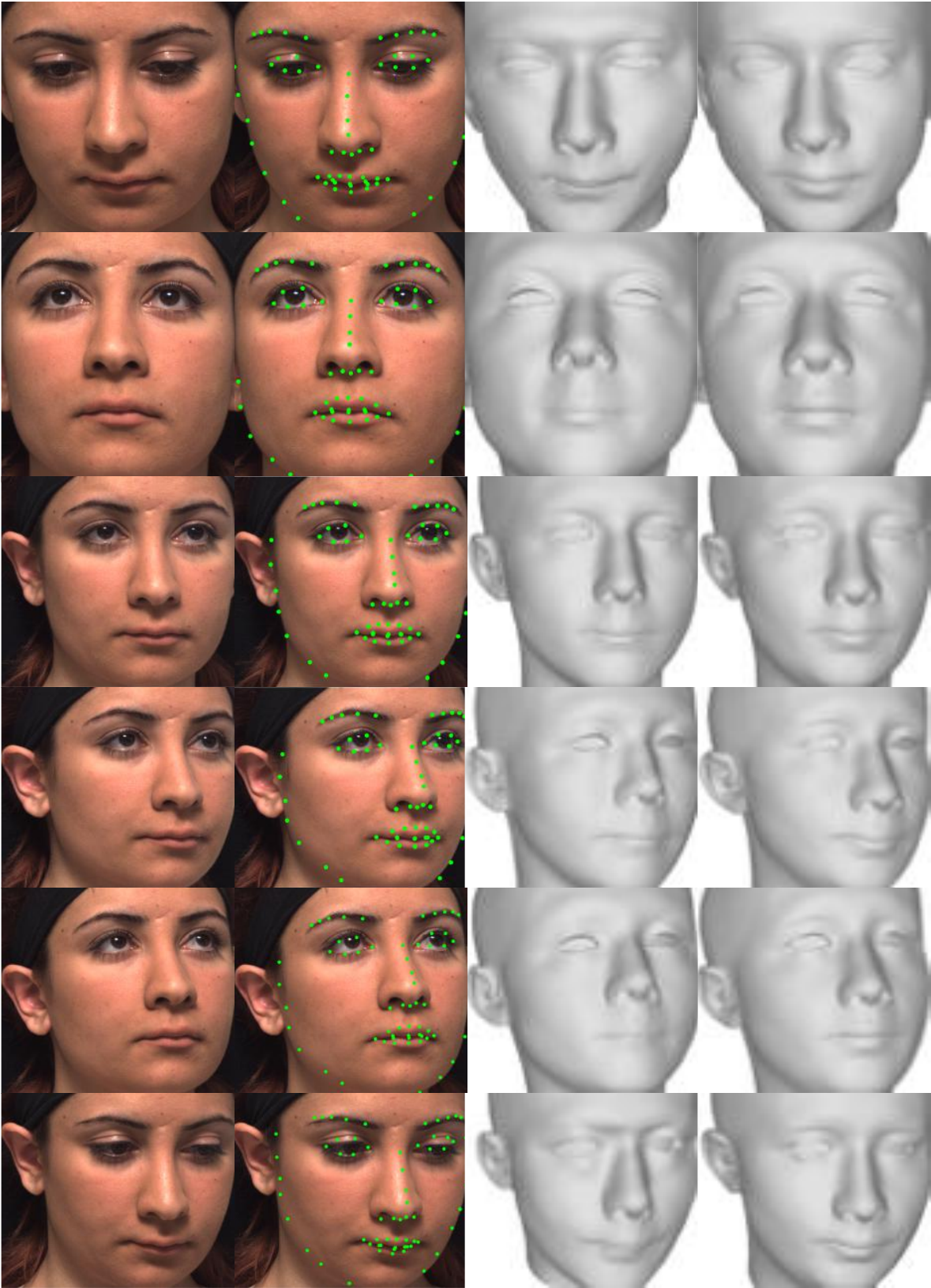}
 \caption{Results of the proposed method over different poses. Col 1: Input image; Col 2: 3D landmark projections; Col 3: Reconstruction results. Col 4: Ground truth.}
 \label{fig6}
\end{figure}
Fig. \ref{fig5} shows the MAE compared with the ground truth mesh $M$ across different poses. It can be observed that the reconstruction accuracy decreases gradually as the yaw rotation angle increases. Furthermore, there are pitch rotations and arbitrary rotations that do not have clear angle labels, and for these samples our method also shows reasonable reconstruction accuracy. Fig. \ref{fig6} illustrates the reconstruction results and 3D landmark projections of different poses. 

\begin{table}[ht]
\renewcommand{\arraystretch}{1.5}
\caption{Comparison with Zhu \MakeLowercase{\textit{et al.}} [6]}
\label{table_1}
\centering
\resizebox{0.5\textwidth}{!}{
\begin{tabular}{|c||c|c|c|c|}
\hline
\multirow{2}{*}{}
 & \multicolumn{2}{|c|}{RMSE with S}&\multicolumn{2}{|c|}{RMSE with M}\\ \cline{2-5}
 & Center&Whole&Center&whole\\
\hline
Zhu \MakeLowercase{\textit{et al.} [6]}&2.7586&3.0641&2.8437&3.8224\\
\hline
Ours&2.4628&2.8522&2.6308&3.5101\\
\hline
\end{tabular}
}
\end{table}
\subsubsection{Comparisons} 
\textit{Zhu \MakeLowercase{\textit{et al.}} on FG2015} We followed the framework of \cite{zhu2015discriminative}, which used cascaded regression based on HOG feature. Compared with \cite{zhu2015discriminative}, we proposed LD feature and combine it with HOG feature, which covers local patch and semantic meaning. Besides, to avoid the limitation of training data, we estimate mapping matrices directly instead of using cascaded regression. Moreover, our method is eligble to deal with different poses and expressions while \cite{zhu2015discriminative} only has application for frontal neutral face. Therefore for fair comparison, only frontal neutral face samples were used for the training and testing in this experiment. Besides, we also showed the comparison with landmark fitting method, which is used as initialization in the proposed method.  
\begin{figure}[ht]
 \centering
  \includegraphics[width=\linewidth]{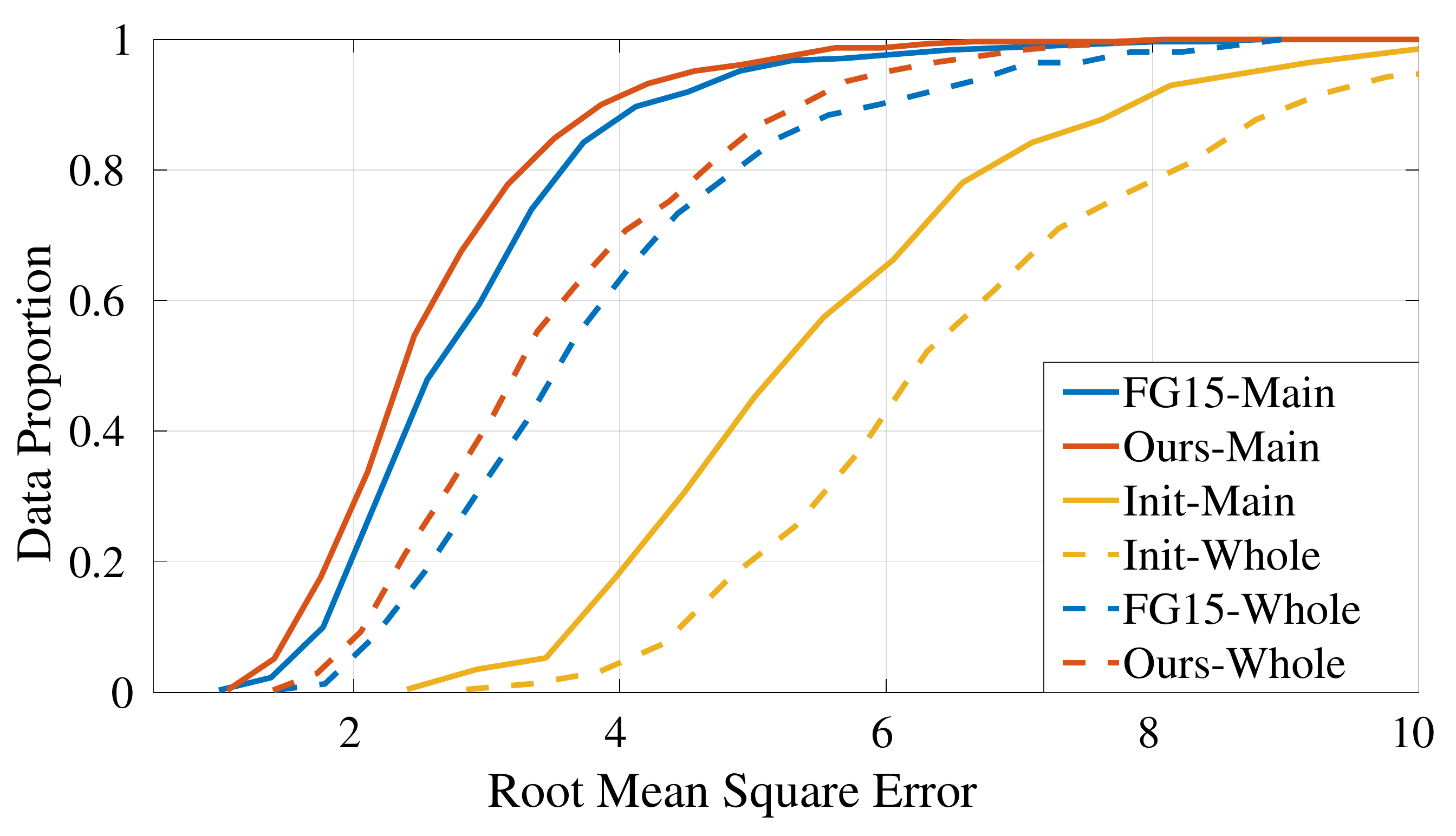}
  \caption{Comparison with Zhu \MakeLowercase{\textit{et al.}}}
  \label{fig4}
\end{figure}
\begin{figure}[ht]
 \centering
  \includegraphics[width=\linewidth]{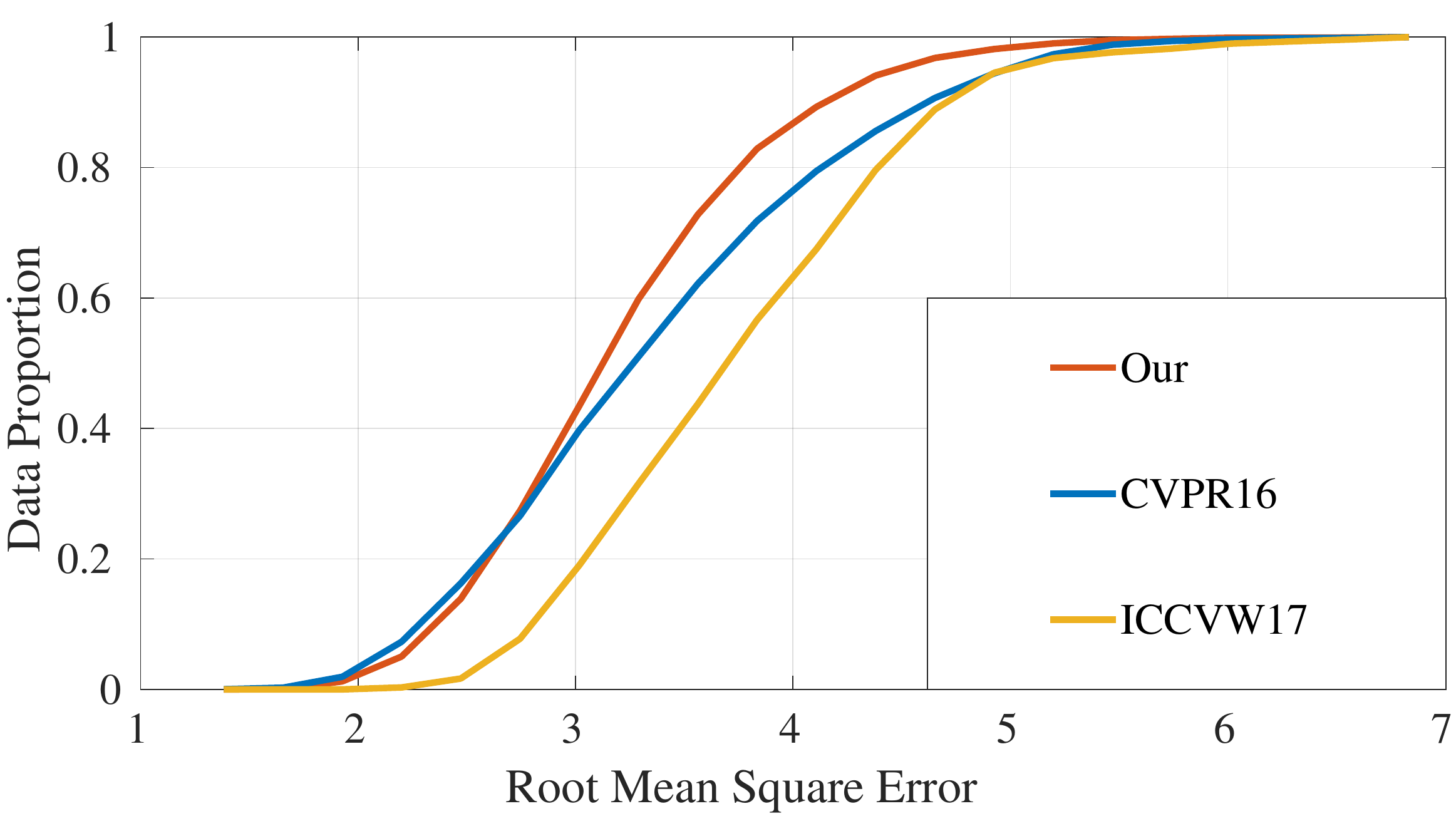}
  \caption{Comparison with Zhu \MakeLowercase{\textit{et al.}} and landmark fitting method}
  \label{fig3}
\end{figure}
Figure \ref{fig4} (a) and (b) show the Cumulative Error Distribution (CED) of the RMSE on the main face and the whole face area, respectively. It can be observed that the proposed method shows a higher reconstruction accuracy than Zhu \MakeLowercase{\textit{at al.}} for both the main face and whole face areas, evaluated with the ground truth mesh. Table \ref{table_1} showed the more detailed comparison with Zhu \MakeLowercase{\textit{at al.}} evaluated with the ground truth model shape $S^\star$ and the ground truth mesh $M$. As mentioned earlier, the ground truth model shape was generated using the 3DMM, while the ground truth mesh was further processed by Laplacian deformation to better fit the input 3D point cloud.
\begin{figure*}
 \centering
 \includegraphics[width=0.8\textwidth]{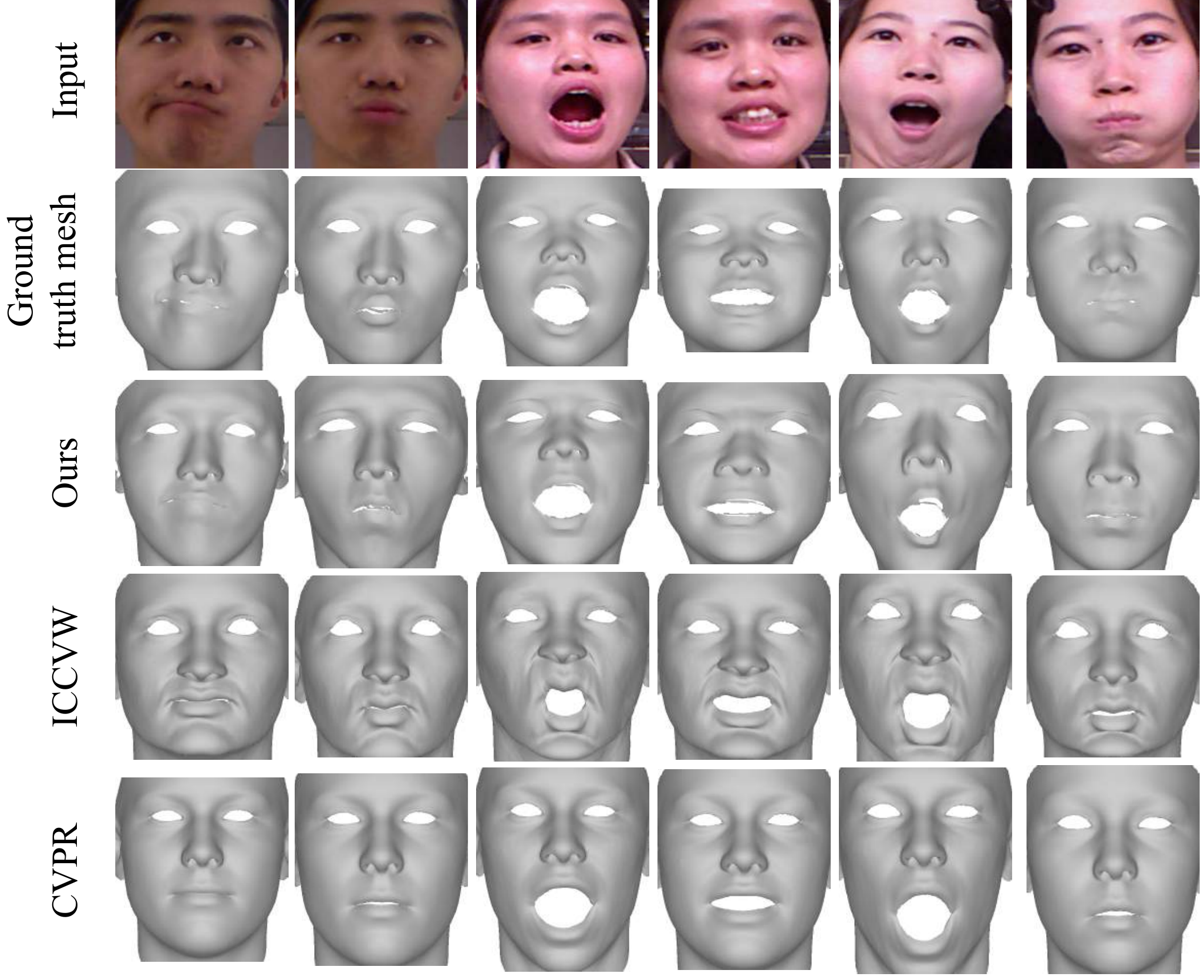}
 \caption{Comparison with \cite{liu2017dense} and \cite{zhu2016face}. Row 1: input images. Row 2: ground truth. Row 3: Ours. Row 4: \cite{liu2017dense}. Row 5: \cite{zhu2016face}}.
 \label{fig10}
\end{figure*}
\par
\textit{Zhu \MakeLowercase{\textit{et al.}} in CVPR2016} \cite{zhu2016face} and \textit{Liu \MakeLowercase{\textit{et al.}} in ICCVW2017} \cite{liu2017dense} We also compared the proposed method with some state-of-the-art methods that used deep neural networks to estimate 3DMM parameters and mapping matrices. In \cite{zhu2016face}, 122,450 face images from 300W-LP \cite{zhu2016face} were used to learn a cascaded Convolution Neural Network(CNN). The residue of 3DMM parameters and mapping matrices are estimated in each stage of the cascade. In \cite{liu2017dense}, 114,848 face images from various datasets were used as the training set. Two CNNs were learned to estimate 3DMM parameters and mapping matrices separately. Compared with them, we only used 4666 face images from the Bosphorus dataset as the training set. On the other hand, we used FaceWarehouse Dataset as the testing set, which contains 3000 images with different identities and expressions. The comparison between the proposed method, \cite{zhu2016face} and \cite{liu2017dense} is showed in Figure \ref{fig3}. Figure \ref{fig10} shows the reconstruction results for samples with different lighting conditions, identities and expressions. It can be observed that our method reconstructs discriminative face shapes for different identities. Moreover, our method can reconstruct extreme expressions, e.g., for the second and sixth samples, our method can reconstruct more accurate mouth shapes.
\par
\begin{figure}[ht]
 \centering
 \includegraphics[width =0.8\linewidth]{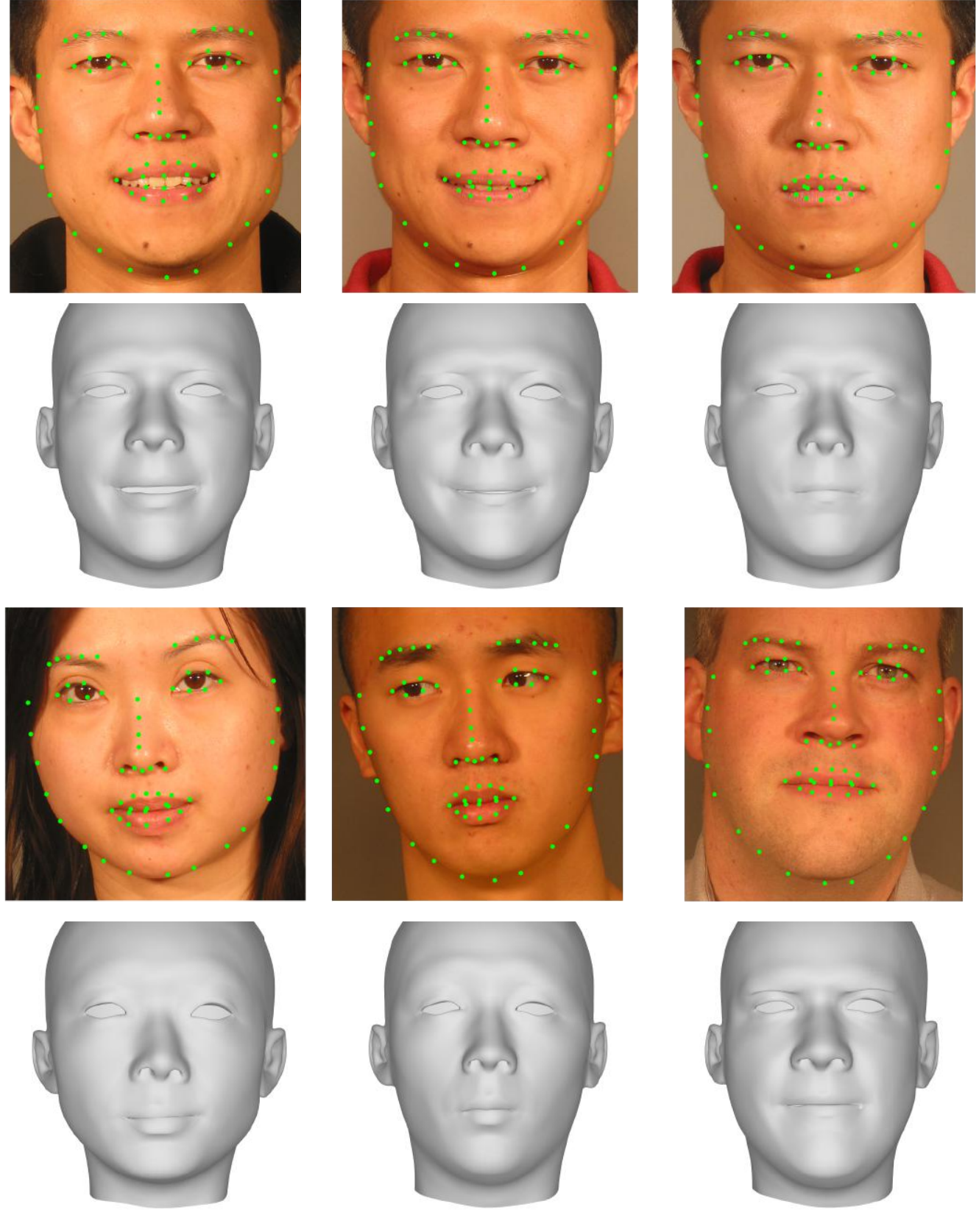}
 \caption{Results on FRGC2 dataset. Row 1 and 3: Input image with 3D landmark projections. Row 2 and 4: Reconstruction results.}
 \label{fig9}
\end{figure}

To further validate the generality of the proposed method, the cascaded regressor learned from the Bosphorus database was tested on another dataset, i.e., Face Recognition Grand Challenge (FRGC2) dataset\cite{frgc2005}. Fig. \ref{fig9} illustrates the 3D reconstruction results and the landmark projections. As shown in the first two rows, the proposed method can reconstruct 3D faces with same identity and different expressions. For different people with similar expressions as shown in the last two rows, our method highlights their difference in terms of the identity and maintains the similar expression. The results on FRGC2 dataset show that our method is robust to different subjects and lighting conditions.

\section{Conclusion}
In this paper, we proposed a 3D face reconstruction method which uses cascaded regression to estimate the 3DMM parameters. Faithful Landmarks detection algorithms are used to provide a sparse correspondence between the 2D image and the 3D face geometry. A joint feature consisting of HOG and LD is used in the regression process, which is robust to lighting, scale and rotation changes. Both subjective and objective evaluations showed the robustness of the proposed method. Compared with the state-of-the-art method, our results achieved lower RMSE in different facial areas. Besides, the proposed method has consistent result in different poses and expressions. The cross-validation on the FaceWarehouse and FRGC datasets proves that the proposed method is robust to lighting variations. In the future, we aim to improve both the quantity and the quality of 3DMM parameters for more 3D face datasets. Besides, 3DMM only provides smooth face shape, as mentioned in Section 1, which could be complemented by SFS methods. Furthermore, as more training data is available, the cascaded regressor will be replaced with other methods such as deep neural network in our future work.

\ifCLASSOPTIONcaptionsoff
  \newpage
\fi


\bibliographystyle{IEEEtran}
\bibliography{fzwubib}
\end{document}